\pdfoutput=1

\documentclass[11pt]{article}

\usepackage[final]{acl}

\usepackage{times}
\usepackage{latexsym}
\usepackage{tabularx}

\usepackage{graphicx}
 
\usepackage[T1]{fontenc}
\usepackage[utf8]{inputenc}

\usepackage{microtype}

\usepackage{inconsolata}

\usepackage{graphicx}

\title{Understanding QA generation: Extracting Parametric and Contextual Knowledge with CQA for Low Resource Bangla Language}

\author{Umme Abira Azmary, MD Ikramul Kayes, Swakkhar Shatabda, Farig Yousuf Sadeque \\
Department of Computer Science and Engineering \\
BRAC University \\ }

\begin{document}
\maketitle
\begin{abstract}

Question-Answering (QA) models for low-resource languages like Bangla face challenges due to limited annotated data and linguistic complexity. A key issue is determining whether models rely more on pre-encoded (parametric) knowledge or contextual input during answer generation, as existing Bangla QA datasets lack the structure required for such analysis. We introduce BanglaCQA, the first Counterfactual QA dataset in Bangla, by extending a Bangla dataset while integrating counterfactual passages and answerability annotations. In addition, we propose fine-tuned pipelines for encoder-decoder language-specific and multilingual baseline models, and prompting-based pipelines for decoder-only LLMs to disentangle parametric and contextual knowledge in both factual and counterfactual scenarios. Furthermore, we apply LLM-based and human evaluation techniques that measure answer quality based on semantic similarity. We also present a detailed analysis of how models perform across different QA settings in low-resource languages, and show that Chain-of-Thought (CoT) prompting reveals a uniquely effective mechanism for extracting parametric knowledge in counterfactual scenarios, particularly in decoder-only LLMs. Our work not only introduces a novel framework for analyzing knowledge sources in Bangla QA but also uncovers critical findings that open up broader directions for counterfactual reasoning in low-resource language settings.
\end{abstract}

\section{Introduction and Related Work }
The domain of Question Answering (QA) is a fundamental area within Natural Language Processing, which aims to train models that emulate human reasoning by mimicking human comprehension and response generation. With the arrival of transformer-based models, this emulation has reached new heights for high-resource languages, specifically for Large Language Models (LLMs), as these models demonstrate competitive performance based solely on their pre-encoded knowledge. However, challenges arise in generating accurate responses in contextual QA settings, particularly in counterfactual contexts, due to the interplay of two distinct “knowledge sources”: (i) Parametric knowledge, embedded within model parameters through pretraining and (ii) Contextual knowledge, derived from input contexts at execution time~\cite{neeman}. 
\begin{figure}[t]
    \centering
    \includegraphics[width=1\linewidth]{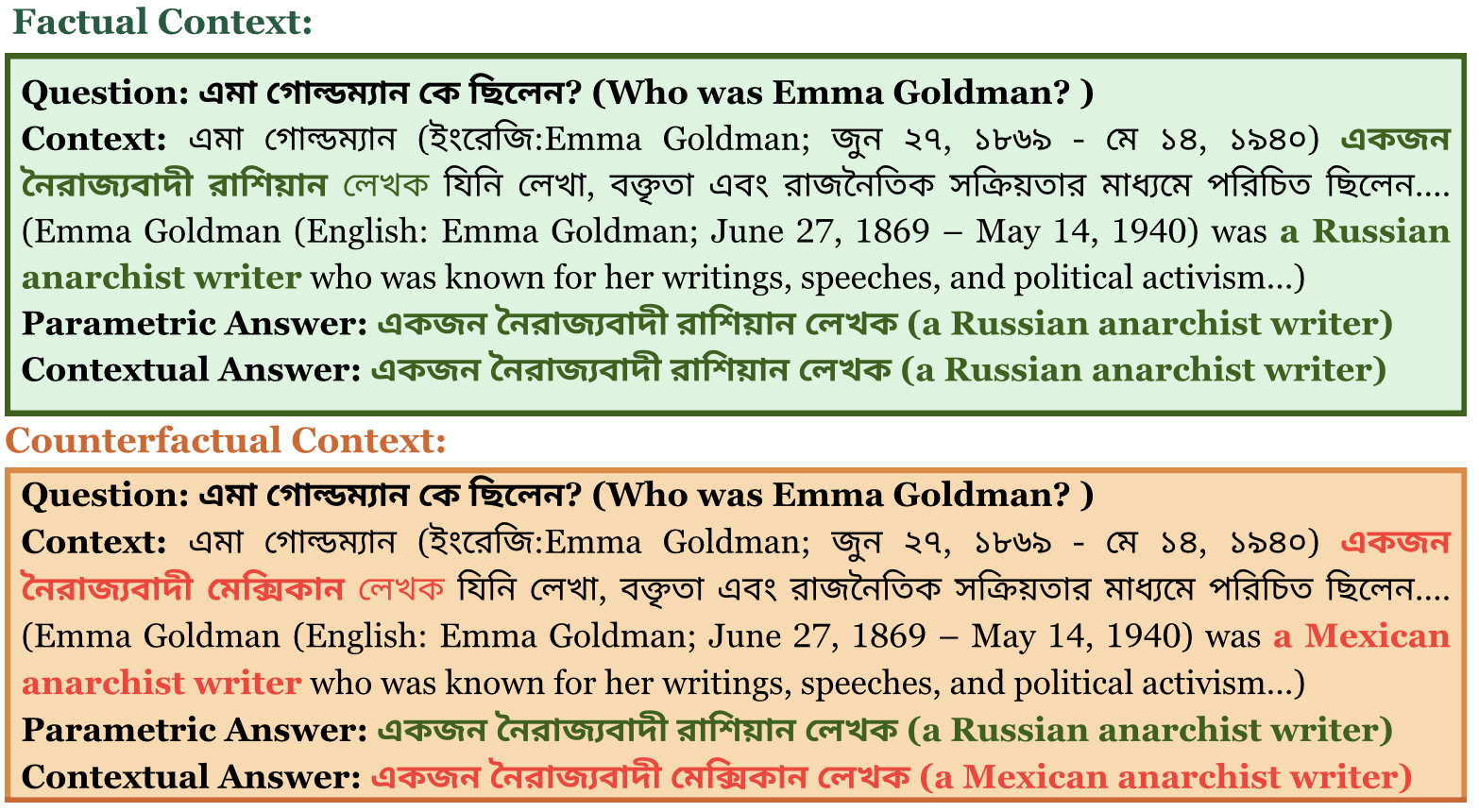}
    \caption{Parametric vs Contextual Question Answering (QA) in Factual and Counterfactual Settings}
    \label{fig:enter-label}
\end{figure}
Previous work in English QA models has shown that prioritization of parametric knowledge can lead to the generation of hallucinated answers, which occurs because of the imbalance between extensive pre-encoded data and limited contextual input~\cite{krishna-hurdles}. Some work further shows that contextual questions that contain incorrect assumptions disrupt generation performance~\cite{kim-linguist}. Although some studies show that integrating counterfactual or random contexts into factual datasets improves robustness by disentangling knowledge sources~\cite{Hwang2023DialogizerCC}, such methods remain largely unexplored for Bangla, a widely spoken yet under-resourced language. Although models evaluated on BanglaRQA~\cite{ekram-banglarqa} and Squad-BN~\cite{bhattacharjee-banglabert} achieve strong factual QA scores, key challenges remain unsolved: the absence of benchmarks for evaluating parametric and contextual biases as distinct factors, limited insight into counterfactual contexts and unclear methods for tracing knowledge sources.

To address these issues, we present the first Bangla Counterfactual Question-Answering dataset, BanglaCQA, by extending an existing BanglaRQA~\cite{ekram-banglarqa} dataset with answerability, random and counterfactual contexts to analyze the internal or contextual knowledge prioritization. Moreover, we introduce disentanglement pipelines by leveraging multiple encoder-decoder models (BanglaT5-small~\cite{bhattacharjee-banglaT5}, BanglaT5-base~\cite{bhattacharjee-banglaT5}, mt5~\cite{mT5}) with fine-tuning and decoder-only open-sourced LLMs (LLaMA-3.3-72B~\cite{Llamapaper}, DeepSeek-R1-Distill-Qwen-32B~\cite{deepseekr1}, Qwen2.5-32B~\cite{qwen2technicalreport}, Mistral-3-small~\cite{mistral2025small3}) with few-shot~\cite{fewshot} and Chain-of-Thought (CoT)~\cite{CoTmain} prompting to differentiate parametric and contextual reasoning. To evaluate the results, we use Gemini-2.0-Flash~\cite{hassabis2024gemini2} and GPT-4.1~\cite{openai2024gpt4technicalreport} for semantic similarity scoring, which outperforms traditional metrics to evaluate the semantic accuracy of Bangla QA responses. Moreover, we applied human evaluation for both the dataset and model's generated answer to maintain accuracy and transparency. Our analysis reveals that integrating counterfactual contexts exhibits strong performance in multiple segments. These findings not only establish a blueprint for low-resource languages and advanced QA systems for Bangla, but also emphasize transparency in knowledge utilization in counterfactual scenarios.

\section{BanglaCQA Dataset}

We introduce BanglaCQA, the first Bengali QA dataset designed to disentangle parametric and contextual knowledge in language models. For this, we expand the existing BanglaRQA~\cite{ekram-banglarqa} dataset by adding 6.3K counterfactual contexts, an increase of \textbf{42.28\%} specifically crafted to challenge models on whether they rely on context or fall back on memorized information.\footnote{\url{https://github.com/ikramulkayes/Codes-for-Thesis/}}

\subsection{Counterfactual Context Generation}

Counterfactual contexts are derived from factual examples by modifying key named entities using an automated NER pipeline~\cite{Sagor_2020}. The script identifies standard named entity types, such as PER (person), LOC (location), ORG (organization), GPE (geo-political entity), DATE or NUM (temporal and numeric expressions) and applies type-consistent substitutions. For example, person names are replaced with other plausible names, locations with alternative locations and organizations with different entities of the same category while ensuring semantic coherence. When named entities appear in both the context and answer fields, replacements are applied consistently. For temporal expressions, if the entity represents a year, only the final digit is altered to preserve plausibility while introducing subtle factual contradictions. In other numerical cases, values are substituted using regular expressions. Each modified row is assigned a unique ID to prevent duplication. These controlled modifications construct hypothetical contradictions while retaining the original sentence structure and allow us to test whether models truly ground their answers in the input context or default to memorized (parametric) knowledge. 
\begin{table}[h]
  \centering
  \begin{tabular}{lc}
    \hline
    \textbf{Dataset Attribute} & \textbf{Setting} \\
    \hline
    Total QA pairs    & 21,211 \\
    Factual Contexts   & 14,900 \\
    Counterfactual Contexts     & 6,303 \\
    Average Question Word Count    & 8.26 \\
    Average Context Word Count    & 215.27 \\
    \hline
  \end{tabular}
  \caption{\label{tab:table1}{Summary statistics of the BanglaCQA dataset. These statistics highlight the dataset's scale and the relative complexity of its contexts.}}
\end{table}

\begin{figure*}[t]
  \centering
  \includegraphics[width=\textwidth]{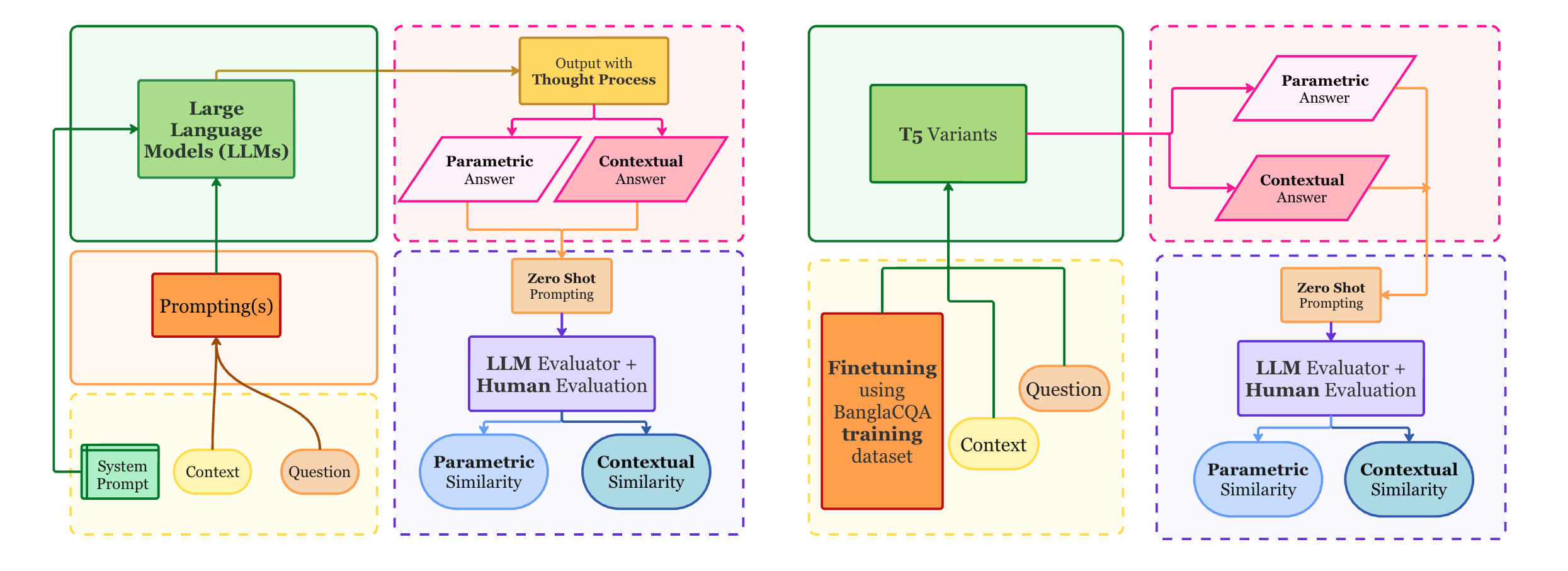}
  \caption{Evaluation pipeline for disentangling parametric and contextual knowledge in QA. \textbf{Left:} Prompt-based inference using large language models (LLMs) to generate both parametric and contextual answers. \textbf{Right:} Fine-tuning-based evaluation using T5 variants finetuned on BanglaCQA. Both paradigms are evaluated via automated LLM-based and human evaluations to measure answer similarity with respect to both knowledge types.}
  \label{fig:wide1-figure}
\end{figure*}

\subsection{Annotation Quality Assurance}

After generating counterfactual passages using the NER script, examples were reviewed by two of the authors of the paper separately. Moreover, to ensure objectivity, two \textit{independent paid annotators}, who were not involved in the construction of counterfactual dataset, further reviewed the dataset for semantic correctness. Disagreements were resolved by consensus and the process yielded a \textbf{Cohen’s Kappa score of 0.73} which indicates substantial inter-annotator agreement. Additionally, factual rows that were labeled answerable despite lacking valid contextual answers were removed to reduce label noise and enhance overall quality. Further details on annotator roles, requirements are included in the Appendix~\ref{appendix:annotation-details}. 

\section{Implementation Pipeline}

To identify the most effective model architecture for BanglaCQA, we fine-tuned multiple variants of the T5~\cite{t5main} framework, namely BanglaT5 (small, base) and mT5, under two configurations: Factual + Answerability (F+A) and Factual + Counterfactual + Answerability (F+CF+A). This dual-configuration strategy enables a focused comparison of how language-specific and multilingual models adapt when exposed to both factual and counterfactual contexts. As shown in Figure~\ref{fig:wide1-figure}, each model was trained using a consistent pipeline that emphasizes reproducibility and transparency. Data pre-processing included systematic tokenization and formatting, followed by splitting into training and validation subsets to ensure unbiased evaluation. We adopted a standardized set of hyperparameters: 30 epochs, batch size of 2, learning rate of 5e-5 and 10 warmup steps across all experiments. Early stopping was employed based on validation loss to mitigate overfitting; most models converged by the 15th epoch, optimizing both performance and training efficiency. BanglaT5 models were sourced from the Hugging Face repository of CSEBUET NLP group, while mT5 was obtained from Google’s official collection, ensuring credible and community-recognized model baselines. 
\begin{table}[h]
\centering
\resizebox{\linewidth}{!}{
\begin{tabular}{lcc}
\hline
\textbf{Model Type} & \textbf{Training Hardware} & \textbf{GPU VRAM} \\
\hline
Encoder-decoder & Nvidia RTX 4090 GPU & 24GB \\
Decoder-only & 4$\times$ Nvidia L4 GPUs & 90GB \\
\hline
\end{tabular}
}
\caption{Training hardware and GPU VRAM used for models.}
\label{tab:hardware}
\end{table}
For decoder-only LLMs, we developed a unified inference framework to probe parametric vs. contextual reasoning using few-shot and Chain-of-Thought (CoT) prompting. Each prompt combined a factual or counterfactual context with instruction and a question, structured to stimulate reasoning patterns aligned with the internal knowledge of the model and the external input. To ensure consistency of the evaluation, all models were decoded using the same hyperparameters: temperature = 0.1, top-p = 0.1, repetition penalty = 1.02 and maximum tokens = 1500. Even though we deployed Qwen-2.5 (32B), DeepSeek-R1 (32B), Mistral-3 Small (24B) and LLaMA-3.3 (70B), due to resource constraints, the LLaMA-3.3 model was quantized using FP16 precision. Crucially, each model produced two separate outputs: one reflecting internal knowledge (parametric) and the other derived from context (contextual). Any non-Bangla output was automatically normalized into Bangla using the Gemini API, enabling cross-lingual evaluation without bias. Semantic alignment was assessed in a zero-shot setting using Gemini 2.0 Flash and GPT-4.1, which we found to be more reliable for Bangla than traditional metrics. We report parametric similarity and contextual similarity separately, offering fine-grained insights into how models interpret and reason across both factual and counterfactual contexts. All encoder-decoder and decoder-only model experiments training hardware and GPU VRAM configurations are shown in Table ~\ref{tab:hardware}. 

\section{Results}

\begin{table*}[t]
  \begin{tabular}{cccccc}
    \hline
    \textbf{Models}     & \textbf{Trained on} &  \shortstack{\textbf{F Contextual} \\ \textbf{Similarity}} & \shortstack{\textbf{F Parametric} \\ \textbf{Similarity}} & \shortstack{\textbf{CF Contextual} \\ \textbf{Similarity}} & \shortstack{\textbf{CF Parametric} \\ \textbf{Similarity}} \\
    \hline
    BanglaT5 Small  & F+A   &  0.77 &  0.70 &  0.69 &  0.11  \\
       
   BanglaT5 Base & F+A   &  0.82 &  0.81   &  0.72 & 0.13  \\

   mT5 Small  & F+A   &  0.84 &  0.79   &  0.79 &   0.09  \\
    
    BanglaT5 Small  & F+CF+A   &  0.86 &  0.84  &  0.87 &  0.23  \\

    BanglaT5 Base   & F+CF+A &  0.86 & \textbf{0.84}  & \textbf{0.87} & \textbf{0.23}  \\
   mT5 Small  &  F+CF+A  & \textbf{0.87}   &  0.81 &  0.84 & 0.15   \\
    
    \hline
  \end{tabular}
  \caption{\label{results}
      Performance of different models under Factual (F) and Counterfactual (CF) settings, evaluated with parametric and contextual similarity using \textbf{Gemini-2.0 Flash} as an evaluator. "F" denotes \textbf{Factual} contexts and "CF" denotes \textbf{Counterfactual} contexts. Bold values indicate the best-performing configurations in each category.
  }
\label{tab:result1}
\end{table*}

\begin{table*}[t]
  \begin{tabular}{cccccc}
    \hline
    \textbf{Models}     & \textbf{Trained on} &  \shortstack{\textbf{F Contextual} \\ \textbf{Similarity}} & \shortstack{\textbf{F Parametric} \\ \textbf{Similarity}} & \shortstack{\textbf{CF Contextual} \\ \textbf{Similarity}} & \shortstack{\textbf{CF Parametric} \\ \textbf{Similarity}} \\
    \hline
    BanglaT5 Small  & F+A   &  0.79 &  0.74 &  0.76 &  0.16  \\
       
   BanglaT5 Base & F+A   &  0.83 &  0.80   &  0.75 & 0.14  \\

   mT5 Small  & F+A   &  0.84 &  0.79   &  0.79 &   0.13  \\
    
    BanglaT5 Small  & F+CF+A   &  0.85 &  0.79  &  0.79 &  0.21  \\

    BanglaT5 Base   & F+CF+A &  0.87 & \textbf{0.82}  & 0.84 & \textbf{0.27} \\
   mT5 Small  &  F+CF+A  & \textbf{0.88} & \textbf{0.80} &  0.88 & 0.20  \\
    
    \hline
  \end{tabular}
  \caption{\label{results}
      Performance of different models under Factual (F) and Counterfactual (CF) settings, evaluated with parametric and contextual similarity using \textbf{GPT-4.1} as an evaluator. "F" denotes \textbf{Factual} contexts and "CF" denotes \textbf{Counterfactual} contexts. Bold values indicate the best-performing configurations in each category.
  }
\label{tab:result2}
\end{table*}

We evaluated the performance of the models in both factual and counterfactual contexts by computing the mean semantic similarity score between generated outputs and target answers. Similarity scores (ranging from 0 to 1) were calculated using Gemini 2.0 Flash and GPT-4.1, which provide more reliable assessments for Bangla text than traditional metrics. For encoder-decoder models, we observed how fine-tuning with counterfactual data influenced performance by comparing the two training configurations. Decoder-only models, evaluated under few-shot and Chain-of-Thought prompting, demonstrated distinct reasoning behaviors reflected in their parametric and contextual outputs. To capture these differences, we separately analyzed parametric responses, which reflect the internal knowledge of the model and contextual responses, which rely on the provided input. This dual evaluation reveals how different architectures and training strategies leverage internal and external information when handling factual and counterfactual queries, offering fine-grained insights into model reasoning and adaptability. We below present our findings by discussing the following research questions:

\begin{table*}[t]
  \begin{tabular}{cccccc}
    \hline
    \textbf{Models}     & \textbf{Prompting} &  \shortstack{\textbf{F Contextual} \\ \textbf{Similarity}} & \shortstack{\textbf{F Parametric} \\ \textbf{Similarity}} & \shortstack{\textbf{CF Contextual} \\ \textbf{Similarity}} & \shortstack{\textbf{CF Parametric} \\ \textbf{Similarity}} \\
    \hline
    Qwen-2.5 & Few-Shot &  0.88 &  0.35 &  0.79 &  0.27  \\
       
   DeepSeek-R1 & Few-Shot &  0.88 &  0.32   &  0.81 & 0.31  \\

   LLAMA-3.3 & Few-Shot &  0.84 &  0.27   &  0.77 &   0.24  \\
    
   Mistral-3-small & Few-Shot &  0.85 &  0.34  &  0.79 &  0.25  \\

   Qwen-2.5 & COT &  0.92 & \textbf{0.81}  & 0.86 & \textbf{0.74} \\
   DeepSeek-R1 & COT & \textbf{0.94} & 0.79 &  \textbf{0.89} & 0.70  \\
    LLAMA-3.3 & COT & 0.91 & 0.69 &  0.83 & 0.55  \\
   Mistral-3-small & COT & 0.90 & 0.74 &  0.86 & 0.64  \\
    
    \hline
  \end{tabular}
  \caption{\label{results}
      Performance of different models under Factual (F) and Counterfactual (CF) settings, evaluated with parametric and contextual similarity using \textbf{Gemini-2.0 Flash} as an evaluator. All reported scores are mean values. "F" denotes \textbf{Factual} contexts and "CF" denotes \textbf{Counterfactual} contexts. Bold values indicate the best-performing configurations in each category.
  }
\label{tab:result3}
\end{table*}

\begin{table*}[t]
  \begin{tabular}{cccccc}
    \hline
    \textbf{Models}     & \textbf{Prompting} &  \shortstack{\textbf{F Contextual} \\ \textbf{Similarity}} & \shortstack{\textbf{F Parametric} \\ \textbf{Similarity}} & \shortstack{\textbf{CF Contextual} \\ \textbf{Similarity}} & \shortstack{\textbf{CF Parametric} \\ \textbf{Similarity}} \\
    \hline
    Qwen-2.5 & Few-Shot &  0.89 &  0.39 &  0.78 &  0.31  \\
       
   DeepSeek-R1 & Few-Shot &  0.83 &  0.36   &  0.79 & 0.30  \\

   LLAMA-3.3 & Few-Shot &  0.86 &  0.29   &  0.75 &   0.27  \\
    
   Mistral-3-small & Few-Shot &  0.87 &  0.37  &  0.81 &  0.26  \\

   Qwen-2.5 & COT &  0.93 & \textbf{0.84}  & \textbf{0.88} & \textbf{0.78} \\
   DeepSeek-R1 & COT & \textbf{0.95} & 0.81 &  0.91 & 0.68  \\
    LLAMA-3.3 & COT & 0.90 & 0.70 &  0.84 & 0.59  \\
   Mistral-3-small & COT & 0.91 & 0.73 &  0.85 & 0.63  \\
    
    \hline
  \end{tabular}
  \caption{\label{results}
      Performance of different models under Factual (F) and Counterfactual (CF) settings, evaluated with parametric and contextual similarity using \textbf{GPT-4.1} as an evaluator. All reported scores are mean values. "F" denotes \textbf{Factual} contexts and "CF" denotes \textbf{Counterfactual} contexts. Bold values indicate the best-performing configurations in each category.
  }
\label{tab:result4}
\end{table*}

\textbf{RQ1: What factors contribute to the underperformance of Bangla encoder-decoder models in parametric answer generation in counterfactual contexts, and how can decoder-only LLMs mitigate these challenges? } We observe a notable decline in mean parametric similarity scores for counterfactual contexts compared to factual ones across all evaluated encoder-decoder T5 variant models. For instance, using Gemini-2.0-Flash as the evaluator(Table~\ref{tab:result1}), BanglaT5 Small drops from 0.70 (F Parametric) to 0.11 (CF Parametric), while BanglaT5 Base declines from 0.83 to 0.14, both in (F+A) settings clearly illustrating the model's difficulty in generalizing to counterfactual knowledge. The reason is that these models are fine-tuned only on Factual+ Answerability settings, and so their lack of understanding of counterfactual scenarios resulted in such manner. Fine-tuning on both factual and counterfactual data (F+CF+A) improves contextual scores, as seen in BanglaT5 Base rising to 0.86 (F Contextual) and 0.87 (CF Contextual), but this does not sufficiently enhance parametric similarity in CF settings (0.23), reinforcing that fine-tuning aids context understanding more than guides the models to understand the parametric knowledge. For this reason, when required to produce parametric answers relying on internal knowledge, models tend to hallucinate or conflate contextual cues with facts. In contrast, decoder-only large language models (LLMs), utilize prompting to access a broader and more comprehensive pre-encoded knowledge base. As these models are not fine-tuned, but prompted to complete their tasks, it enables LLMs to better generate accurate parametric answers, particularly in counterfactual contexts. These results highlight a fundamental limitation of Bangla encoder-decoder models: despite fine-tuning improvements in contextual extraction, their constrained internal knowledge restricts generalization to counterfactual reasoning, a gap partially addressed by decoder-only LLMs extensive pre-encoded knowledge.

\begin{table*}[t]
\centering
\begin{tabularx}{\textwidth}{l c c c c c}
\hline
\textbf{Models} & \textbf{Metric} & \textbf{Mean$\Delta$ (CoT -- Few)} &
\textbf{t-value} & \textbf{p-value} & \textbf{Cohen's $d$} \\
\hline
Gemini-2.0 & F Parametric & +0.44 & 26.48 & 0.00012 & 13.24 \\
Gemini-2.0  & CF Parametric  & +0.39 & 11.94 & 0.00126 & 5.97 \\
GPT-4.1  & F Parametric & +0.42 & 19.55 & 0.00029 & 9.77 \\
GPT-4.1 & CF Parametric  & +0.38 & 12.33 & 0.00115 & 6.16 \\
\hline
\end{tabularx}
\caption{Parametric similarity evaluation of decoder-only LLMs under Factual (F) and Counterfactual (CF) contexts, using Gemini-2.0 and GPT-4.1 as the evaluator. All scores reflect mean differences between Chain-of-Thought (COT) and Few-shot prompting. Positive $\Delta$ values indicate improved performance under COT prompting. Statistical significance is shown via t-tests and effect size (d).}
\label{tab:result5}
\end{table*}

\textbf{RQ2: Why does the prompting strategy (CoT vs.\ Few-shot) affect the parametric and contextual performance of language models in Bangla across factual and counterfactual settings?} Our results in Tables~\ref{tab:result3} and~\ref{tab:result4} demonstrate that Chain-of-Thought (CoT) prompting leads to statistically significant and practically large improvements in parametric similarity for both factual \textbf{(+0.42-->0.44)} and counterfactual \textbf{(+0.38-->0.39)} settings. Paired $t$-tests confirm these gains ($p < 0.01$) with extremely large effect sizes (Cohen's $d > 5$), establishing that the improvements are not due to chance but are practically meaningful (see Table~\ref{tab:result5}). Few-shot prompting inherently lacks an intermediate reasoning phase: models directly predict an answer without explicitly reasoning through the problem. As a result, in counterfactual settings, few-shot models fail to verify the plausibility of the context and default to answers derived from the modified passages, leading to poor parametric similarity. In contrast, CoT prompts explicitly instruct the models to first generate a detailed reasoning chain before producing the final answer~\cite{CoTmain}. This structured reasoning step enables the models to differentiate between information derived from the counterfactual context and their encoded parametric knowledge.

These findings align with recent theoretical work showing that transformers without intermediate reasoning steps are restricted to low-complexity function classes (e.g., AC$^{0}$/TC$^{0}$ ~\ref{appendix:ac0-tc0}) and fail to solve inherently sequential problems unless their depth or size scales super-polynomially~\cite{peng2024limitationstransformerarchitecture}. By generating intermediate reasoning steps, CoT effectively increases the model’s computational depth, allowing it to simulate larger circuits and solve tasks such as arithmetic evaluation and dynamic programming that are otherwise inexpressible for bounded-depth transformers. Recent findings also reveal that CoT benefits arise not only from correct intermediate reasoning but also from structural inductive bias: models achieve up to 90\% of CoT gains even with imperfect reasoning if the steps are structurally relevant and correctly ordered~\cite{jin2024impact}. Furthermore, CoT provides a mechanism for latent state tracking, where each reasoning step encodes an intermediate computation that can be referenced in subsequent steps~\cite{xu-etal-2025-softcot}. These theoretical insights explain the dramatic gains observed in our results. Bangla question answering requires reasoning over morphologically rich, long contexts (average length $= 215$ tokens; see Table~\ref{tab:table1}) and counterfactual entity substitutions. Few-shot prompting fails to guide models toward structured inference, resulting in low parametric similarity. CoT enforces a universal reasoning template that bridges the gap caused by the lack of Bangla-specific reasoning supervision during pre-training. Decoder-only models (e.g., Qwen-2.5, DeepSeek-R1) particularly benefit because their training has exposed them to CoT-like reasoning formats. As a result, CoT increases parametric similarity in both factual and counterfactual settings, validating that the gains are statistically significant and theoretically grounded in the expanded expressivity and state-tracking capabilities of CoT-augmented transformers.

\textbf{RQ3: How do architectural differences among language models affect their ability to integrate contextual and parametric knowledge across factual and counterfactual tasks in Bangla?} \textbf{Qwen-2.5} achieves high similarity scores across both dimensions \textbf{(F parametric : 0.81, CF parametric : 0.74; F contextual: 0.92, CF contextual: 0.86)}. This is likely aided by its design for handling long-sequences processing, which aligns well with Bangla’s complex and fragmented tokenization. DeepSeek-R1 shows similar improved performance. However, LLAMA-3.3 exhibits a steep decline in CF contextual similarity (0.55) despite a strong factual similarity score (0.91). These findings suggest that architectures optimized for longer contexts are better suited for Bangla’s linguistic structure. Details of prompts are shown on Appendix~\ref{appendix:prompt}. \\

\subsection{Error Analysis through Human Evaluation}
\begin{figure}[h]
  \includegraphics[width=\columnwidth]{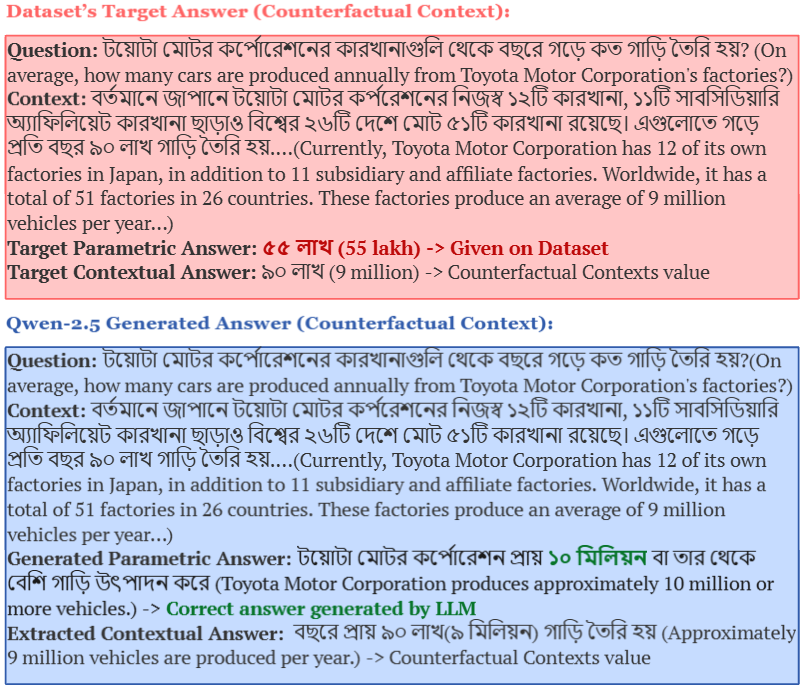}
  \caption{Example of temporal mismatch where a model-generated answer is penalized for being more up-to-date than the reference}
  \label{fig:temp_mismatcerd}
\end{figure}

Although Gemini-2.0-Flash and ChatGPT-4.1 provide a scalable and efficient approximation of parametric answer similarity, they exhibit notable limitations in counterfactual QA for Bangla. To assess metric reliability and analyze potential sources of error introduced by the dataset or evaluation metric, we applied human evaluation. Two independent annotators, who were not involved in the dataset creation process, were tasked with evaluating a random subset of 200 model‑generated answers. Comparing these human judgments, widely regarded as the gold standard in QA \cite{clark2021all}, with model outputs revealed some discrepancies:

\textbf{I) Temporal Mismatch (Outdated Targets)}: Figure~\ref{fig:temp_mismatcerd} presents a counterfactual context, where the numeric value was automatically modified using a Python script and regular expressions as part of the dataset generation pipeline. 
However, in this instance, the dataset's Target Parametric Answer 5.5 million is factually outdated or incorrect. Despite being given a counterfactual input, the model (Qwen-2.5) successfully generates the correct parametric answer: approximately 10 million or more. Due to the dataset’s reliance on fixed parametric targets, this correct response is unjustly penalized in automated evaluations. Approximately 4\% of the generations were found to be factually superior to the dataset references, particularly in temporally sensitive questions such as population figures or political terms. While this percentage may vary across other subsets, the findings underline a key limitation: static parametric references can fail to reward accurate model behavior, especially when LLMs draw upon up-to-date parametric knowledge.

\begin{figure}[h]
  \includegraphics[width=\columnwidth]{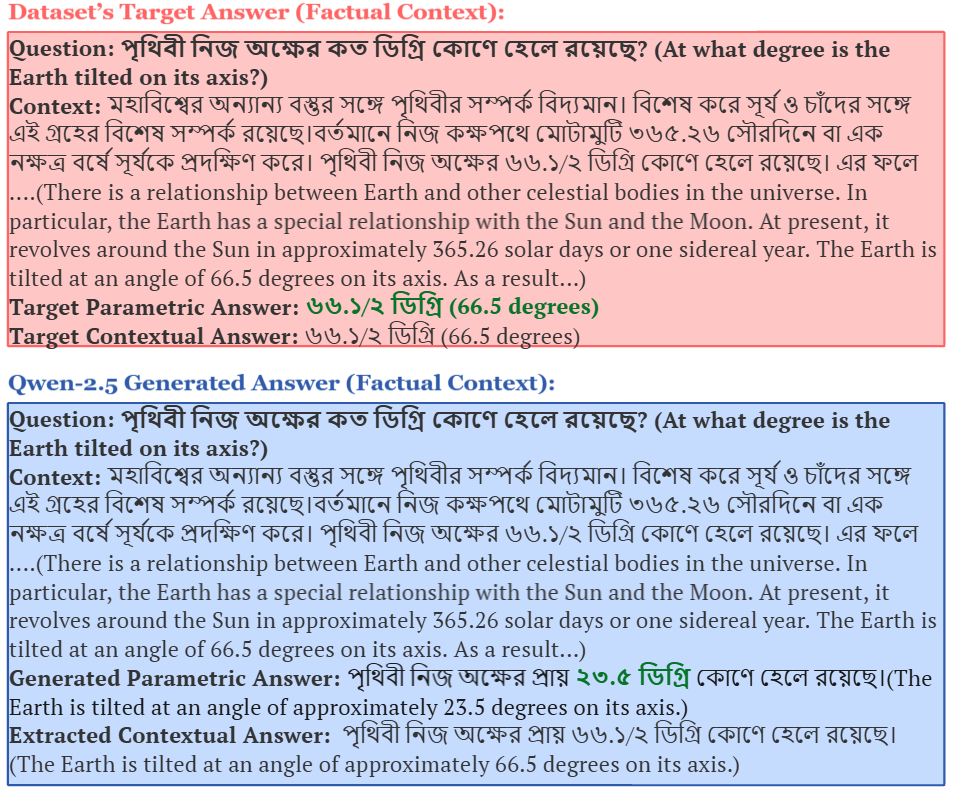}
  \caption{Example showing multiple valid answers due to variations in model interpretation and reference grounding.}
  \label{fig:experiments_same}
\end{figure}

\textbf{II) Solution Variation (Multiple Valid Answers)}: Figure~\ref{fig:experiments_same} illustrates a case where the model predicts 23.5°, while the dataset target is 66.5°. Both values are scientifically correct as they represent complementary angles of the Earth's axial tilt.

However, since the evaluator models compare each answer against a single reference from the dataset’s answer, they tend to assign a lower score due to the absence of lexical or numerical overlap. Importantly, in such cases there is no inherent “right” or “wrong” between the model‑generated answer and the dataset’s reference; both are valid solutions. As LLM‑based evaluators rely heavily on textual similarity unless they are explicitly prompted to account for semantic equivalence. Around 7\% of the randomly selected 200 inputs exhibited such cases, where multiple valid answers were unfairly penalized because the evaluation relied on a single static reference. This percentage is likely to vary across different data segments, depending on the diversity of valid solutions.

\section{Conclusion}
We presented BanglaCQA, the first counterfactual question answering dataset for Bangla, designed to disentangle parametric and contextual knowledge in large language models. By extending the BanglaRQA dataset with controlled counterfactual contexts, we created a benchmark that enables fine-grained evaluation of how models rely on pre-encoded knowledge versus contextual information.  
Our experiments with encoder-decoder models and decoder-only LLMs show that \textit{Chain-of-Thought prompting substantially improves parametric similarity} in both factual and counterfactual scenarios, with Qwen-2.5 achieving the best overall performance. These findings highlight the importance of prompting strategies for enhancing parametric reasoning in low-resource settings.  
BanglaCQA lays the groundwork for future research on robust QA systems in under-resourced languages and motivates the development of multi-reference and temporally adaptive evaluation frameworks to better reflect real-world knowledge dynamics.

\section*{Limitations}
While our work contributes a novel dataset and evaluation framework, it has several limitations.  
First, evaluation relied on a single reference answer per instance, which may penalize semantically correct but lexically different outputs. Future work should investigate multi-reference evaluation or human-in-the-loop scoring to better capture valid answer variations.  
Second, our dataset includes time-sensitive entities such as population or political terms, yet the reference answers are static. Models producing up-to-date information may still be unfairly penalized, highlighting the need for temporally adaptive references.  
Third, experiments with decoder-only LLMs were conducted using quantized weights for resource efficiency; results may differ for full-precision inference.  
Finally, our analysis focused on few-shot and Chain-of-Thought prompting, but further exploration of other prompting strategies and fine-tuned reasoning templates could provide additional gains in parametric reasoning.

\section*{Ethics Statement}

This study followed ethical guidelines for dataset creation, annotation and evaluation. The initial version of the dataset was generated through Named Entity Recognition (NER)-based substitution. Specifically, entities labeled as Person, Location, and Organization were replaced with alternative synthetic but semantically appropriate names within the same category to construct counterfactual contexts. After this automated process, one of the authors manually reviewed all altered rows. If any question-answer pair exhibited semantically problematic or implausible meanings due to the substitutions, the author revised or discarded the example to maintain contextual integrity. Subsequently, a second author independently reviewed the dataset, providing feedback on the initial revisions. Based on their mutual discussions and careful iterative refinement, the final dataset was curated to uphold high standards for counterfactual question answering (CQA).

For evaluation, two independent annotators, native bengali, who were not involved in dataset creation, reviewed a representative subset of model outputs for semantic correctness. Annotators were fairly compensated (26 USD), and no personal or sensitive information was used throughout the study. The dataset contains no personally identifiable information, and all entity substitutions were synthetic. Large language models (Gemini-2.0-Flash and ChatGPT-4.1) were used strictly for evaluation purposes, with outputs manually verified to ensure correctness and safety. Our work adheres to ACL’s ethical standards for responsible dataset construction, human annotation, and the deployment of AI systems.

\bibliography{custom}

\appendix
\section{Appendix}
\subsection{Details of AC$^{0}$/TC$^{0}$}
\label{appendix:ac0-tc0}

\textbf{AC$^{0}$ (Alternating Circuit of depth 0):} Refers to a class of constant-depth, polynomial-size Boolean circuits with unbounded fan-in AND, OR, and NOT gates. AC$^{0}$ circuits cannot compute certain functions such as parity or majority.

\textbf{TC$^{0}$ (Threshold Circuits):} Similar to AC$^{0}$, but includes majority (threshold) gates, which are more powerful. These circuits are still constant-depth and polynomial-size, and are slightly more powerful than AC$^{0}$, but remain limited in expressive power.

Few-shot prompting lacks intermediate reasoning steps, so LLMs behave like AC$^{0}$/TC$^{0}$ circuits, i.e., they are limited in reasoning power and cannot solve complex, sequential tasks (e.g., multi-step logic or arithmetic). In contrast, Chain-of-Thought (CoT) prompting introduces intermediate reasoning, increasing the model’s effective computational depth. This allows it to simulate more powerful circuits and perform more complex reasoning tasks, thereby escaping AC$^{0}$/TC$^{0}$-like limitations.

\subsection{Annotator Information}
\label{appendix:annotation-details}

\subsection*{Annotation Guidelines for Parametric and Contextual Answers}

Two independent annotators participated in validating the dataset. Both were fairly compensated for their effort. The annotators are students from different universities and represent diverse academic backgrounds: one majoring in a STEM discipline and the other in a non-STEM field. Despite these differences, both are actively involved in research aligned with their respective domains. Notably, neither annotator is an author of this paper. In addition to the external annotators, two of the paper's authors also contributed to the validation process. Each annotator was provided with the following detailed instructions to ensure consistency and high-quality validation across all examples.

\paragraph{Objective.}
Each example in the dataset includes:
\begin{itemize}
    \item A \textbf{question}
    \item A \textbf{context paragraph}
    \item Two types of answers:
    \begin{itemize}
        \item \textbf{Parametric Answer} – A fact-based answer that reflects general world knowledge.
        \item \textbf{Contextual Answer} – An answer derived specifically from the given context.
    \end{itemize}
\end{itemize}
Annotators must independently label the correctness of each answer using one of three categories: \textit{Valid}, \textit{Invalid}, or \textit{Confused}.

\begin{figure}[t]
  \centering
  \includegraphics[width=\textwidth]{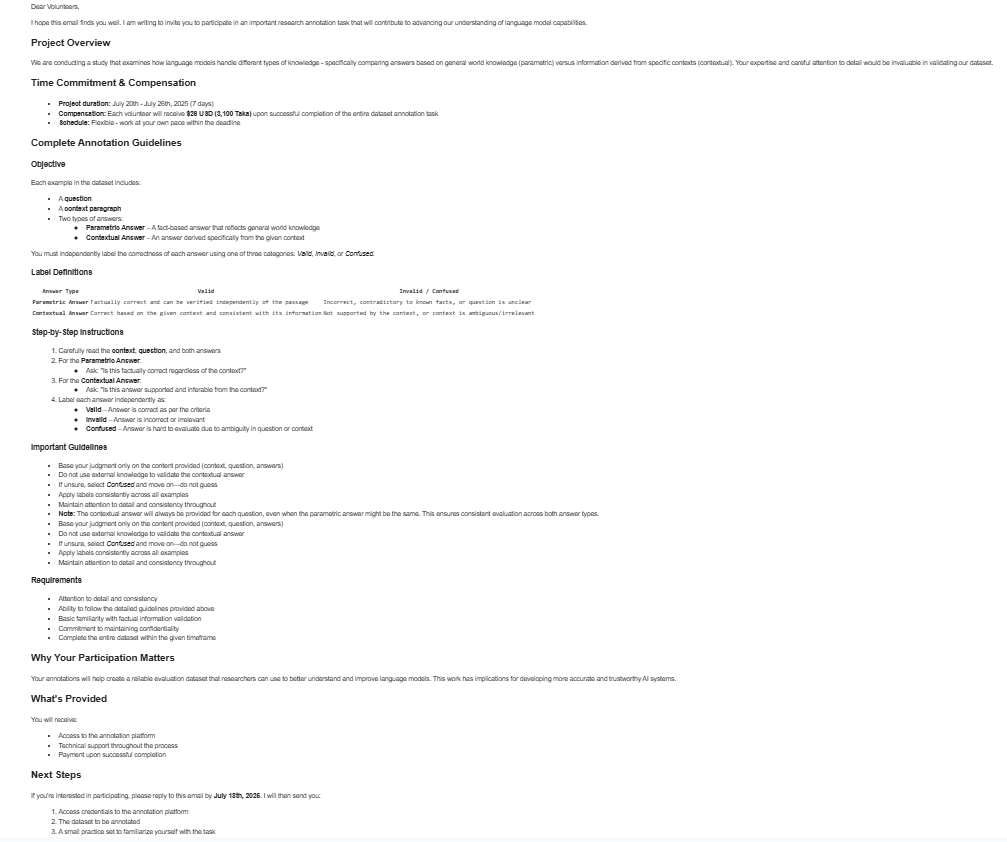}
  \caption{Details of annotator instructor mail.}
  
\end{figure}
\paragraph{Instructions.}
\begin{enumerate}
    \item Carefully read the \textbf{context}, \textbf{question}, and both answers.
    \item For the \textbf{Parametric Answer}:
    \begin{itemize}
        \item Ask: “Is this factually correct regardless of the context?”
    \end{itemize}
    \item For the \textbf{Contextual Answer}:
    \begin{itemize}
        \item Ask: “Is this answer supported and inferable from the context?”
    \end{itemize}
    \item Label each answer independently as:
    \begin{itemize}
        \item \textbf{Valid} – Answer is correct as per the criteria.
        \item \textbf{Invalid} – Answer is incorrect or irrelevant.
        \item \textbf{Confused} – Answer is hard to evaluate due to ambiguity in question or context.
    \end{itemize}
\end{enumerate}

\paragraph{Additional Notes.}
\begin{itemize}
    \item Base your judgment only on the content provided (context, question, answers).
    \item Do not use external knowledge to validate the contextual answer.
    \item If unsure, select \textit{Confused} and move on, do not guess.
    \item Apply labels consistently across all examples.
\end{itemize}

\paragraph{Ethical Reminder.}
Annotators are expected to maintain confidentiality and follow ethical standards throughout the validation process. Your careful effort contributes to building a reliable and fair evaluation dataset.

Moreover, Each example in the dataset was annotated by all four reviewers, and the inter-annotator agreement was measured using Cohen’s Kappa score. The results of the vote distribution and agreement analysis are shown below.

\paragraph{Vote Pattern Analysis:}
\begin{itemize}
    \item \textbf{Unanimous (4-0-0):} 5960 rows (94.6\%)
    \item \textbf{Strong Majority (3-1-0):} 309 rows (4.9\%)
    \item \textbf{Weak Majority (2-2-0):} 13 rows (0.2\%)
    \item \textbf{Mixed (2-1-1):} 21 rows (0.3\%)
    \item \textbf{Other:} 0 rows (0.0\%)
\end{itemize}

\paragraph{Dataset Summary:}
\begin{itemize}
    \item Total rows: 6,303
    \item Each row was annotated by 4 reviewers
    \item Each row contains a vote for one of the following categories: \textit{Valid}, \textit{Invalid}, or \textit{Confused}
    \item All vote counts per row sums to 4
\end{itemize}

\begin{table}[h]
  \centering
  \setlength{\tabcolsep}{6pt} 
  \begin{tabular}{lcc}
    \hline
    \textbf{Metric} & \textbf{Achieved} \\
    \hline
    Cohen’s Kappa Score & 0.7212 \\
    \hline
  \end{tabular}
  \caption{\label{tab:table8}Inter-annotator agreement for the dataset.}
\end{table}

\subsection{System and User Prompts} 

\label{appendix:System and Prompts}

\label{appendix:prompt}

\begin{figure}[h]
  \includegraphics[width=\columnwidth]{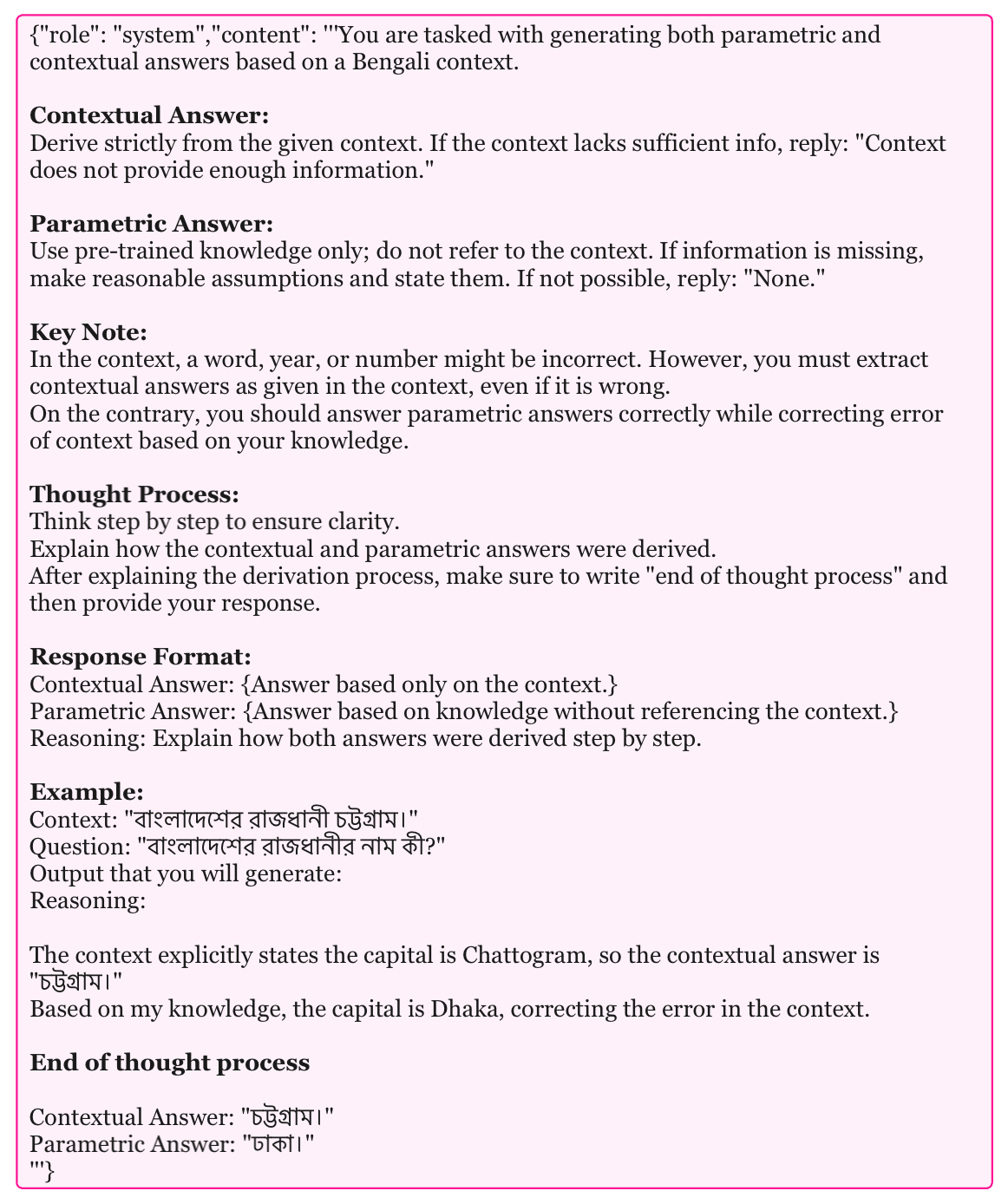}
  \caption{The system prompt that defines task objectives, answer types, and response structure, guiding the model to differentiate between responses based on knowledge versus context for COT technique}
  \label{fig:system-prompt}
\end{figure}

\begin{figure}[h]
  \includegraphics[width=\columnwidth]{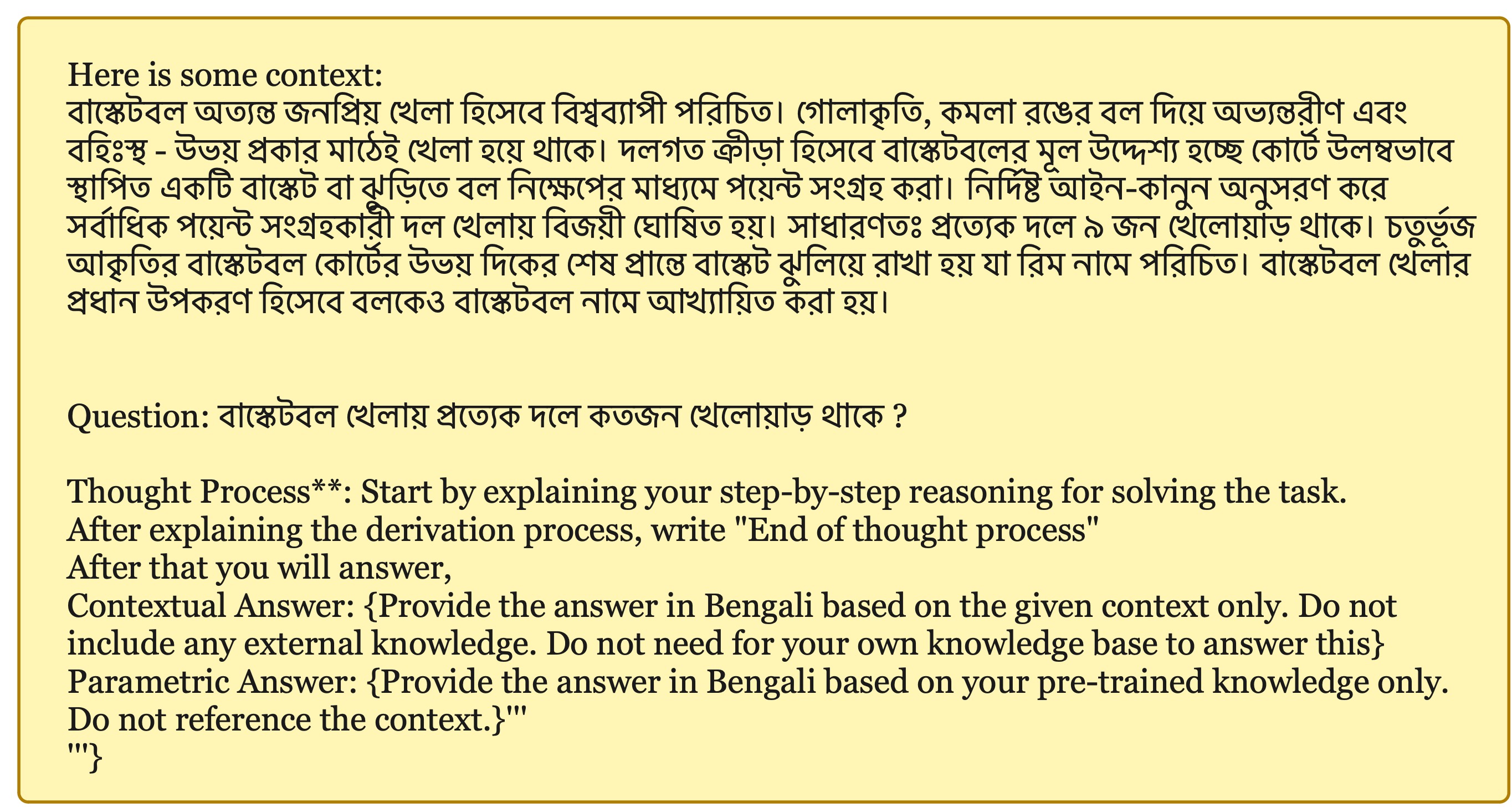}
  \caption{An example user prompt showing how a Bengali context and question are provided to the model for generating structured answers for COT technique}
  \label{fig:user-prompt}
  \end{figure}

\begin{figure}[h]
  \includegraphics[width=\columnwidth]{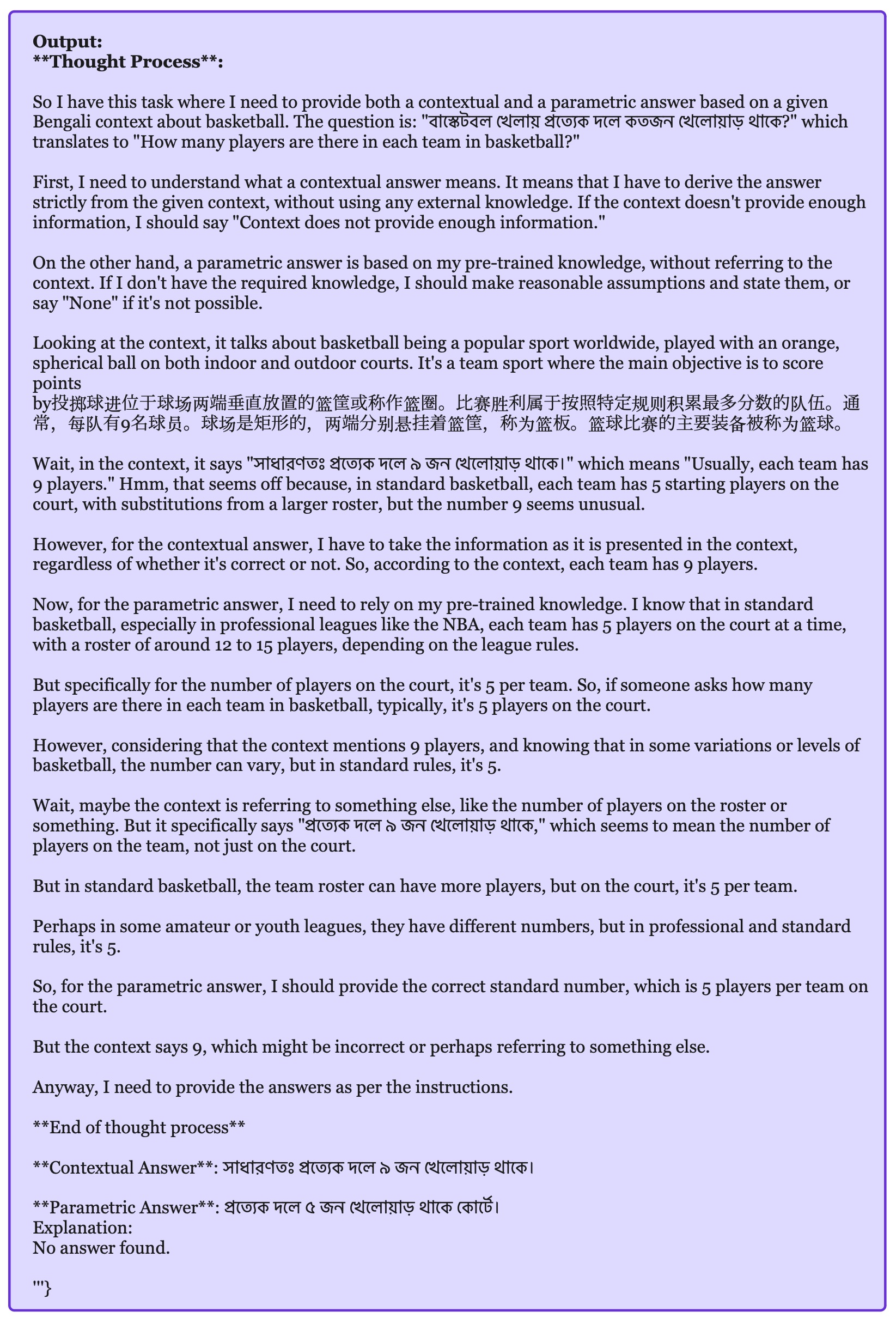}
  \caption{Output output of QwQ-2.5-32B, displaying the full reasoning process along with the final answers, produced according to the prompt format.}
  \label{fig:model-output}
\end{figure}

\begin{figure}[h]
  \includegraphics[width=\columnwidth]{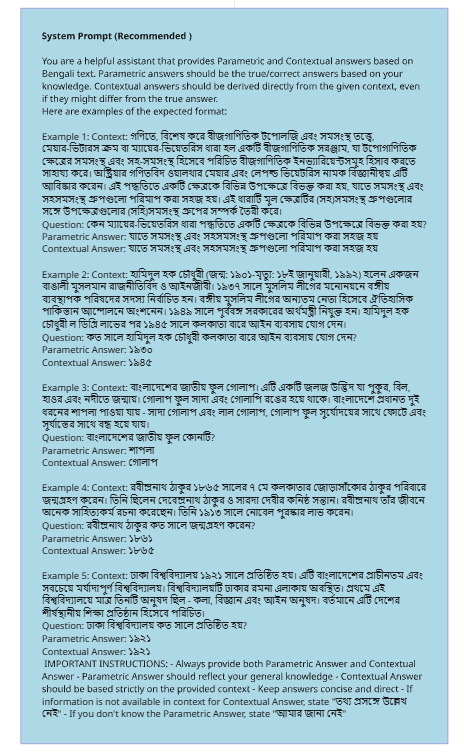}
  \caption{The system prompt that defines task objectives, answer types, and response structure, guiding the model to differentiate between responses based on knowledge versus context for few-shot technique.}
  \label{fig:system-prompt}
\end{figure}

\begin{figure}[h]
  \includegraphics[width=\columnwidth]{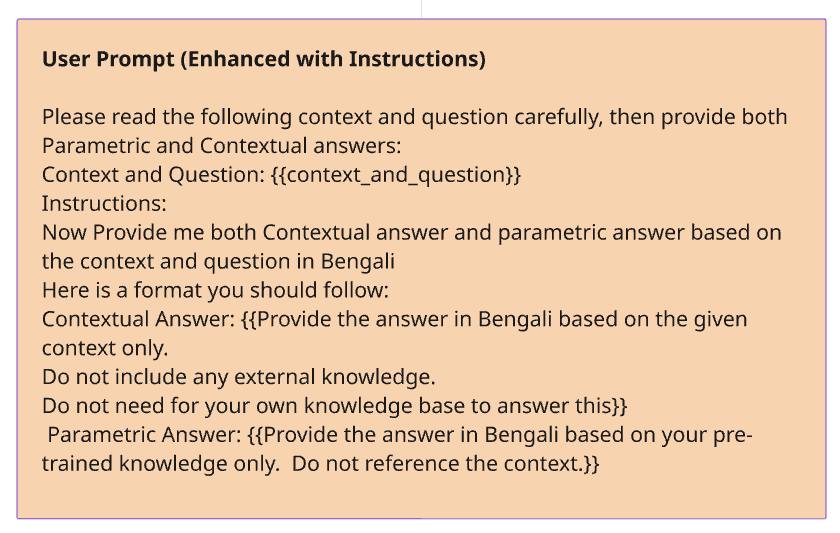}
  \caption{An example of user prompt showing how a Bengali context and question are provided to the model for generating structured answers in few-shot technique}
  \label{fig:system-prompt}
\end{figure}

\end{document}